\documentclass[conference]{IEEEtran}

\ifCLASSINFOpdf
  \usepackage[pdftex]{graphicx}
  \DeclareGraphicsExtensions{.pdf,.jpeg,.png}
\else
\fi

\usepackage{amsfonts, amsmath, amsbsy, amssymb,  epsfig, epstopdf}
\usepackage{nopageno}
\usepackage[all]{xy}

\newcommand{\Fc}{{\mathcal F}}

\newcommand{\Xc}{{\mathcal X}}
\newcommand{\Yc}{{\mathcal Y}}

\begin{document}
%
\title{Wavelet Domain Residual Network (WavResNet) for Low-Dose X-ray CT Reconstruction}

\author{
\IEEEauthorblockN{Eunhee Kang}
\IEEEauthorblockA{
KAIST \\
Daejeon, Korea\\
Email: keh0220@kaist.ac.kr}
\and
\IEEEauthorblockN{Junhong Min$^{\dagger}$}
\IEEEauthorblockA{Samsung Electronics Co.\\
Suwon, Korea\\
Email: gminimok@gmail.com}
\and
\IEEEauthorblockN{Jong Chul Ye$^*$}
\IEEEauthorblockA{
KAIST\\
Daejeon, Korea\\
Email: jong.ye@kaist.ac.kr}
}

\maketitle

\begin{abstract}

Model based iterative reconstruction (MBIR) algorithms for low-dose X-ray CT are computationally complex because of the repeated use of the forward and backward projection.
Inspired by this success of deep learning in computer vision applications, we  recently proposed a deep convolutional neural network (CNN) for low-dose X-ray CT 
and won the second place in 2016 AAPM Low-Dose CT Grand Challenge.
However, some of  the texture are not fully recovered, which was unfamiliar to the radiologists.
To cope with this problem,  here we propose a direct residual learning approach on directional wavelet domain
to solve this problem and to  improve the performance against previous work.
In particular, the new network estimates the noise of each input wavelet transform, and then the de-noised wavelet coefficients are obtained by subtracting the noise from the input wavelet transform bands. 
The experimental results confirm that the proposed network has significantly improved  performance, preserving the detail texture of the original images.

\end{abstract}

\IEEEpeerreviewmaketitle

\section{Introduction}
\label{sec:introduction}

Due to the risk of radiation exposure, methods for minimizing  X-ray dose have been extensively studied. 
A reduction in the number of X-ray photons emitted can solve the problem of radiation exposure.
However, it brings the low signal-to-ratio (SNR) measurements that cause the noise in the reconstruction results.
The noise of low-dose CT is usually approximated by Gaussian model, 
but CT specific streaking noise is also included in the low-dose CT images.
Streaking noise occurs due to the photon starvation and beam hardening 
from sophisticated non-linear X-ray photon acquisition processes.
To cope with these problems, model based iterative reconstruction (MBIR) algorithms have been investigated \cite{beister2012iterative,ramani2012splitting}. 
However, MBIR have the limitations  because of computationally extensive iterative applications of forward and backward projection.

Nowadays, extensive data is available, so it is desirable to use this database.
In the computer vision community, deep convolution neural network (CNN) were actively investigated using the large data and high-performance graphical processing units (GPUs)  
\cite{krizhevsky2012imagenet} .
With the developments of new network units such as rectified linear unit(ReLU), 
max pooling and batch normalization, 
the classical training problems are solved and  the networks are given  deep structures.
Deep network has achieved the great performance improvement in low-level computer vision applications such as denoising \cite{chen2015learning} 
and super-resolution \cite{dong2014learning}.

In the area of medical imaging, there are also extensive research activities that use deep learning. However, most of these studies focus on image-based diagnostics, and its applications for image reconstruction problems such as x-ray computed tomography (CT)  reconstruction are relatively less studied.
Recently, we have introduced a wavelet domain deep learning  algorithm for low-dose X-ray CT algorithm \cite{kang2016deep}, whose validity has been rigorously confirmed by winning the second place award in  AAPM Low-Dose CT Grand Challenges.
However, in this earlier work,  the reconstruction results lost some texture of the original images.
Therefore, one of the most important contributions of this paper is the development of a drastically improved deep network, which overcomes the limitations of previous work by maintaining  detailed textures and edges
to improve the performance.

The key to such an improvement is the observation that the low-dose noise artifacts in wavelet domain has a much simpler topology than the original full-dose images so that the learning of the artifact signal
is easier than learning the full-dose images. Once the noise in the wavelet domain is estimated, the denoised wavelet coefficients are obtained by subtracting the estimated noise from the wavelet coefficients of input
 low-dose X-ray CT images. Then, the final image is obtained by executing wavelet recomposition. 
 Because the learning is done to estimate the wavelet domain residual signals, we call the new deep learning algorithm as {\em wavelet domain residual network (WavResNet)}.
 
%
%

\begin{figure*}[!t]
\centering{\includegraphics[width=16cm]{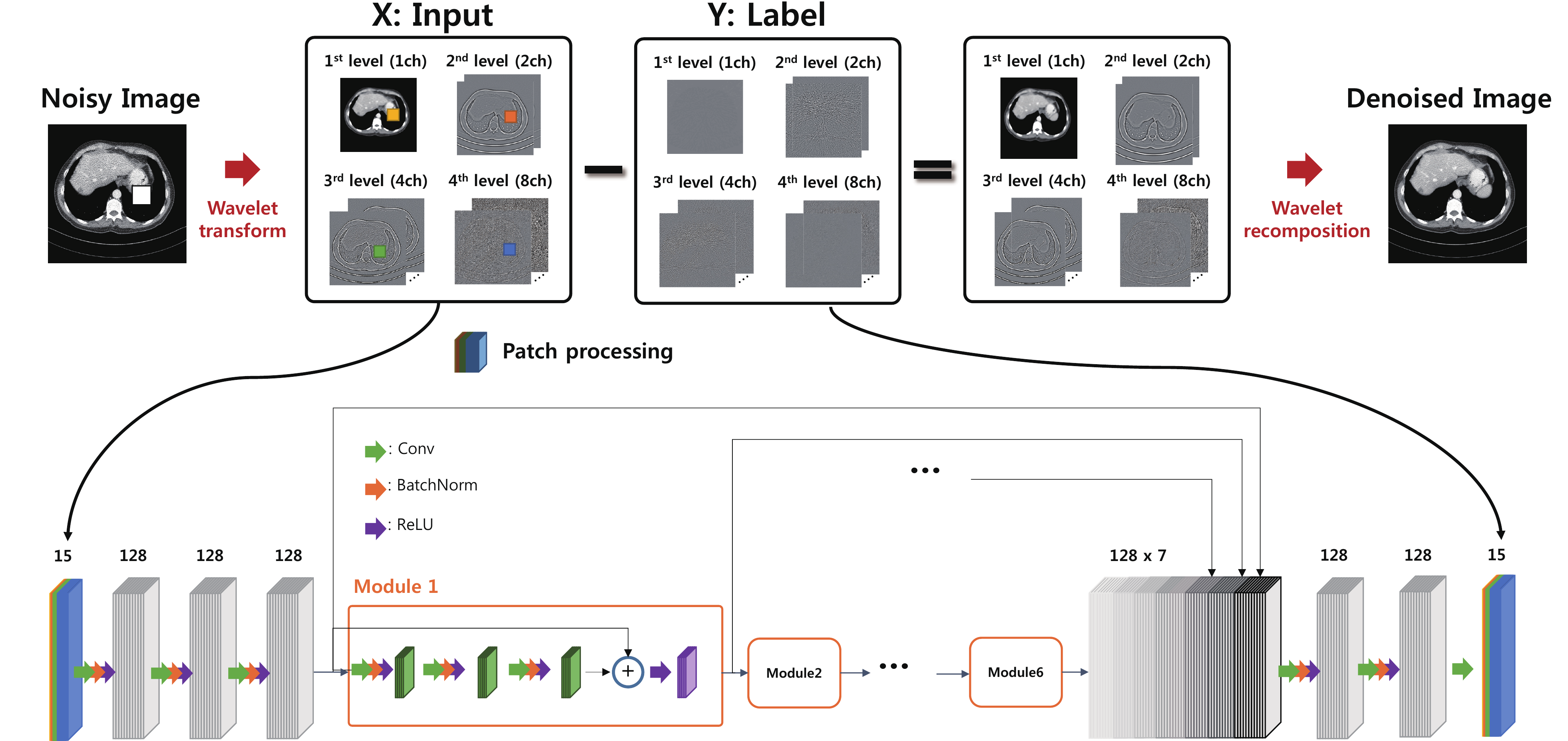}}
\caption{The proposed WavResNet architecture for low-dose X-ray CT reconstruction.}
\label{fig:network_architecture}
\end{figure*}

\section{Theory}
\label{sec:background}

\subsection{Deep learning in higher dimensional feature space}
In a learning problem, based on a observation ({\em input})
$X \in \Xc$ and a {\em label}  $Y \in \Yc$ generated by a distribution $D$, we are interested in  estimating a  regression function $f: X \rightarrow Y$ in a functional space $\Fc$
that minimizes the  risk
$L(f) = E_D \|Y-f(X)\|^2.$ 
However, an important technical issue 
 is that the associated probability distribution $D$ is unknown.
Moreover, we only have a finite
 sequence of training data set
$S=\{(X_1,Y_1),\cdots, (X_n, Y_n)\}$, so there is 
only 
 an empirical risk
$\hat L_n(f) = \frac{1}{n}\sum_{i=1}^n \|Y_i - f(X_i)\|^2.$
A direct minimization of empirical risk is, however,  problematic because of the  overfitting.

To solve this problem, statistical learning theory  \cite{anthony2009neural} has been developed to 
bound the risk of a learning algorithm in terms of complexity  (eg. VC dimension, shatter coefficients, etc) and the empirical
risk. Rademacher complexity \cite{bartlett2002rademacher} is one of the most modern notions of complexity, which is distribution-dependent
and defined for any class of real-valued functions.
Specifically, with probability $\geq 1-\delta$,  for every  function $f\in \Fc$, 
\begin{equation}\label{eq:L}
L(f) \leq \underbrace{\hat L_n(f)}_{\text{empirical risk}} + \underbrace{2 \hat R_n(\Fc)}_{\text{complexity penalty}}+ 3 \sqrt{\frac{\ln(2/\delta)}{n}}
\end{equation}
where the empirical Rademacher complexity $\hat R_n(\Fc)$ is defined to be
$$\hat R_n(\Fc)=  E_\sigma \left[\sup_{f\in \Fc} \left(\frac{1}{n}\sum_{i=1}^n \sigma_i f(X_i) \right) \right],$$ 
where $\sigma_1,\cdots, \sigma_n$ are independent random variables that are uniformly chosen  from $\{-1,1\}$.

Here, empirical risk is determined by the representation power of the network \cite{telgarsky2016benefits}, 
while the complexity is determined by the structure of a network.
The capacity of functions grows exponentially with respect to the number of hidden units \cite{telgarsky2016benefits}. 
When the network architecture is determined, its capacity is fixed. 
Therefore, the performance of the network is now dependent on the complexity of the label $Y$ that a given deep network tries to approximate.

One of the most important contributions of this paper is to extend this idea to a novel deep network design principle.
Specifically, for a given deep network $f: X\rightarrow Y$,
our design goal is to find $T$ and $S$ maps that embed the input and label data sets to high dimensional features space.
Thus, 
the resulting data sets  $X'=T(X)$ and $Y'=S(Y)$ may have simpler data manifold.
This can be shown in the following diagram:
\begin{eqnarray*}
\xymatrix@C+1em@R+1em{
   X \ar[r]^{g}
    \ar@<-2pt>[d]_{T} & Y \ar@<-2pt>[d]_{S} \\
   X'\ar@<-2pt>[u]_{T^{\dag}} \ar[r]^{f} & Y' \ar@<-2pt>[u]_{S^{\dag}}
}
\end{eqnarray*}
where  the superscript $^\dag$ denotes the pseduo-inverse that restore the original input data from high-dimensional features.
Then, it is our goal  to find a neural network $f: X' \rightarrow Y'$   in the feature space that is executed better than in the original image space.
Note that this idea resembles the {\em kernel trick}  \cite{smola1998learning} in   support vector machines, which designs a linear classifier in high dimensional
feature space using nonlinear kernels.

We are aware that the basic motivation of deep learning is to automatically extract the embedding of the  feature space  by learning; thus,  additional embedding
by $T$ and $S$ appear redundant. Although  true in theory,  in many of the medical image reconstruction problems, the database is  usually not
large enough. Thus,  analytic form of the embedding at the input and output ends provides many practical advantages and improved performance.
For example,  in  recent deep  residual learning \cite{zhang2016beyond}, the input transform  $T$ is an identity mapping
and the label transform is given by
\begin{eqnarray*}
Y'= S(Y) = Y-X \quad  ,
\end{eqnarray*}
which we call a {\em  residual transform.}
In other word,  the output embedding is to provide the residual between the artifact-free output and noisy input so that the learning is performed to learn the residual.
Using persistent homology analysis,  our recent work \cite{han2016deep} showed that the label manifold of the residual  data is topologically simpler than that of $Y$.

\subsection{Proposed embedding scheme}

Inspired by this finding,  this paper proposes a high dimensional embedding scheme to  directional wavelet transform domain residual signals.
More specifically,   the wavelet transform  can annihilate the 
 smoothly varying signals while maintaining the image edges  due to the vanishing  moments of wavelets, resulting in the dimensional reduction and
 manifold simplification.
 Furthermore, low-dose X-ray CT images exhibits streaking noise.
Therefore, directional wavelet transform, such as contourlet transform, has directional filter banks that are good in detecting the streaking noise patterns 
and  the directional edge information of X-ray CT images. In addition, the contourlet transform is a redundant transform,  so the associated embedding to
 higher dimensional features space is good for learning.
Therefore, by combining these with the observation that the residual signals have a simpler topological structures,  our proposed embedding $T$ for the input space is contourlet transform
and the $S$ for the label is the residual transform in the contourlet domain, i.e. 
\begin{eqnarray*}
X'=T(X)  &=& \mbox{Contourlet Transform}(X)\\
 Y'=S(Y) &= & T(Y)-T(X).
\end{eqnarray*}
The resulting WavResNet structure is illustrated in Fig.~\ref{fig:network_architecture}.


\section{Method}
\label{sec:method}

\subsection{Network architecture}

In detail, the non-subsampled contourlet transform  consists of two steps \cite{zhou2005nonsubsampled}; 
non-decimated multi-scale decomposition and  directional decomposition.  as shown  in upper part of the Fig. \ref{fig:network_architecture}.
The filter bank does not have down-sampling or up-sampling, so  it is shift invariant.
We used the four level decomposition resulting in 15 bands.

The proposed deep network is shown in lower part of the Fig. \ref{fig:network_architecture}.
It contains 24 convolution layers, followed by a batch normalization layer and a ReLU layer except the last convolution layer.
At the first convolution layer, 128 set of $3\times3\times15$ convolution kernels are used.
Then, 128 set of $3\times3\times128$ convolution kernels are used in the following convolution layers.
The module has 3 sets of convolution, batch normalization, and ReLU layers, and 1 bypass connection has a  ReLU layer.
Out deep network consists of 6 modules and has a channel concatenation layer that stacks several inputs of the individual modules.
This allows gradients to be back-propagated over different paths.

\subsection{Network training}

We trained the proposed deep network by the conventional error back-propagation with stochastic gradient descent (SGD) scheme.
The learning rate was initially set to 0.01 and decreased   continuously down to $10^{-5}$.
We used the gradient clipping method in the range $[-10^{-3}, 10^3]$ to use a high learning rate in the initial training steps.
The size of mini-batch was 10,  and the size of patch was $55\times55$  for training.
The proposed deep network was implemented using MatConvNet \cite{vedaldi2015matconvnet} in MATLAB 2015a enviroment (Mathwork, Natick).

\subsection{Training dataset}

We produced the image data from projection data provided by ``2016 Low-Dose CT Grand Challenge".
The given raw projection data was obtained by a 2D cylindrical detector and a helical conebeam trajectory using a z-flying focal spot. 
These have been  converted to conventional fanbeam projection data by a single slice rebinning technique. 
CT images were reconstructed by using a conventional filtered backprojection (FBP) algorithm.
The training datasets are composed of routine dose and quarter dose X-ray CT images of 10 patients.
The size of the X-ray CT image is $512\times512$ and the slice thickness is 3mm.
We used contourlet transform to routine dose X-ray CT images and quarter dose X-ray images.
Wavelet coefficients from a quater dose were used for the input $X'$
and the difference wavelet coefficients between from quarter dose and routine dose image were used for the label $Y'$.

\subsection{Baseline algorithms}

For a quantitative evaluation, one patient's data was used  and  various image metrics 
such as peak signal-to-noise ratio (PSNR), structural similarity (SSIM) index, and normalized root mean square error (NRMSE) values were calculated.
The test images were denoised by the proposed method and other algorithms such as total variation-based MBIR
 and 
the previous CNN in the `2016 Low-Dose CT Grand Challenge \cite{kang2016deep}, which we call the AAPM-Net.
MBIR regularized by TV was solved by an alternative direction method of multiplier (ADMM) techniques \cite{ramani2012splitting} and Chambolle's proximal TV \cite{chambolle2004algorithm}.
The parameter was selected by numerical experiments.
Compared to WavResNet in Fig.~\ref{fig:network_architecture},  the AAPM-Net directly learns the countourlet coefficients (with the exception of the lowest frequency residuals) and
the other CNN structure is identical.
In addition, these two networks were trained with the 9 training datasets  of the 10  datasets,
and the remaining one data set was used for validation.

\section{Experimental results}
\label{sec:result}

\subsection{Improvement over AAPM-Net}

\begin{figure}[!hbt]
\centering{\includegraphics[width=9cm]{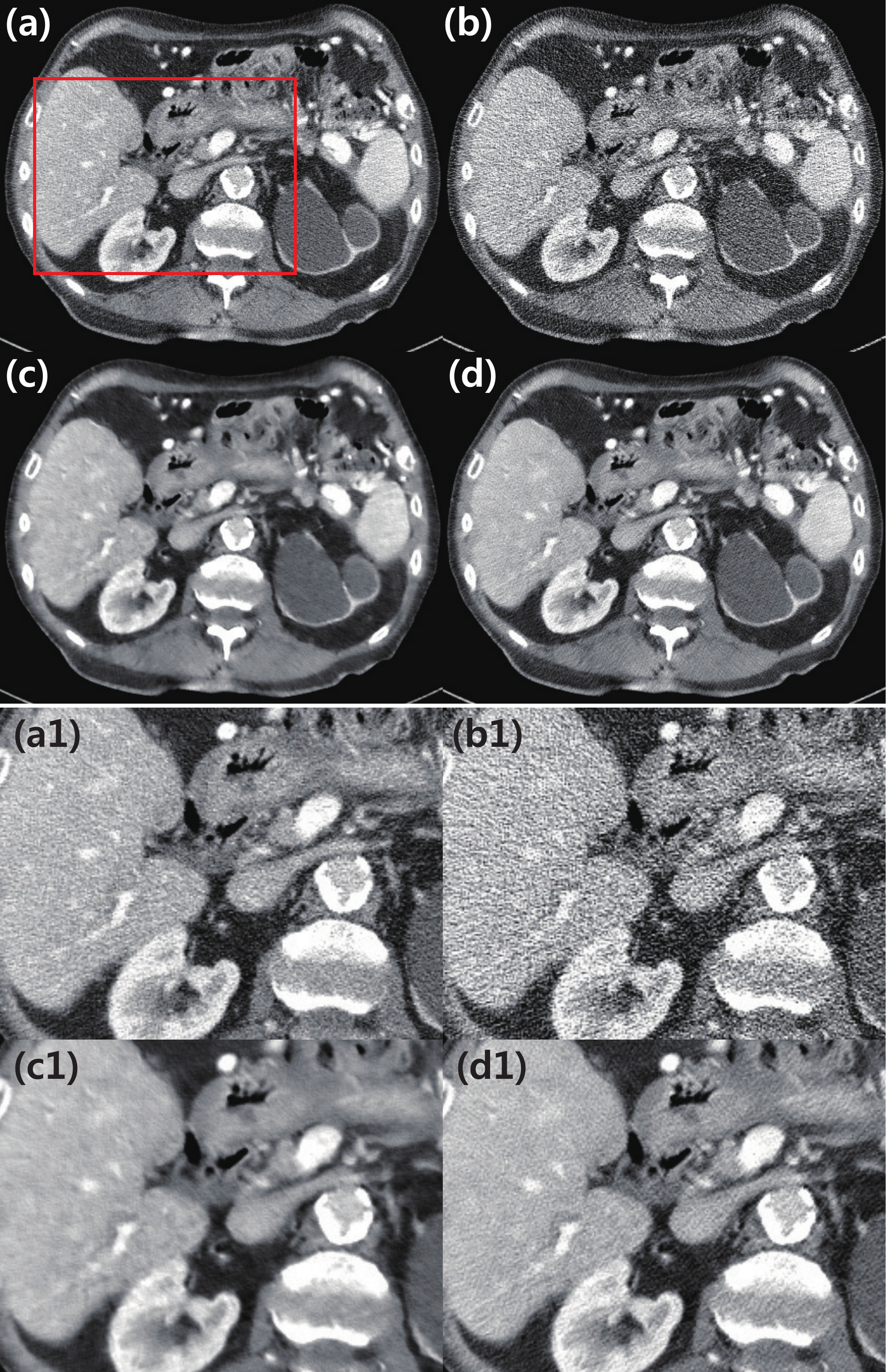}}
\caption{(a-d) Denoised images with routine dose and quarter dose images.
(a1-d1) Magnified images inside the red box.
The intensity range was set to (-160, 240) [HU] (Hounsfield Unit).
(a,a1) Routine dose, (b,b1) Quarter dose,
(c,c1) AAPM-Net, (d,d1) WavResNet.}
\label{fig:result}
\end{figure}

\begin{figure}[!hbt]
\centering{\includegraphics[width=9cm]{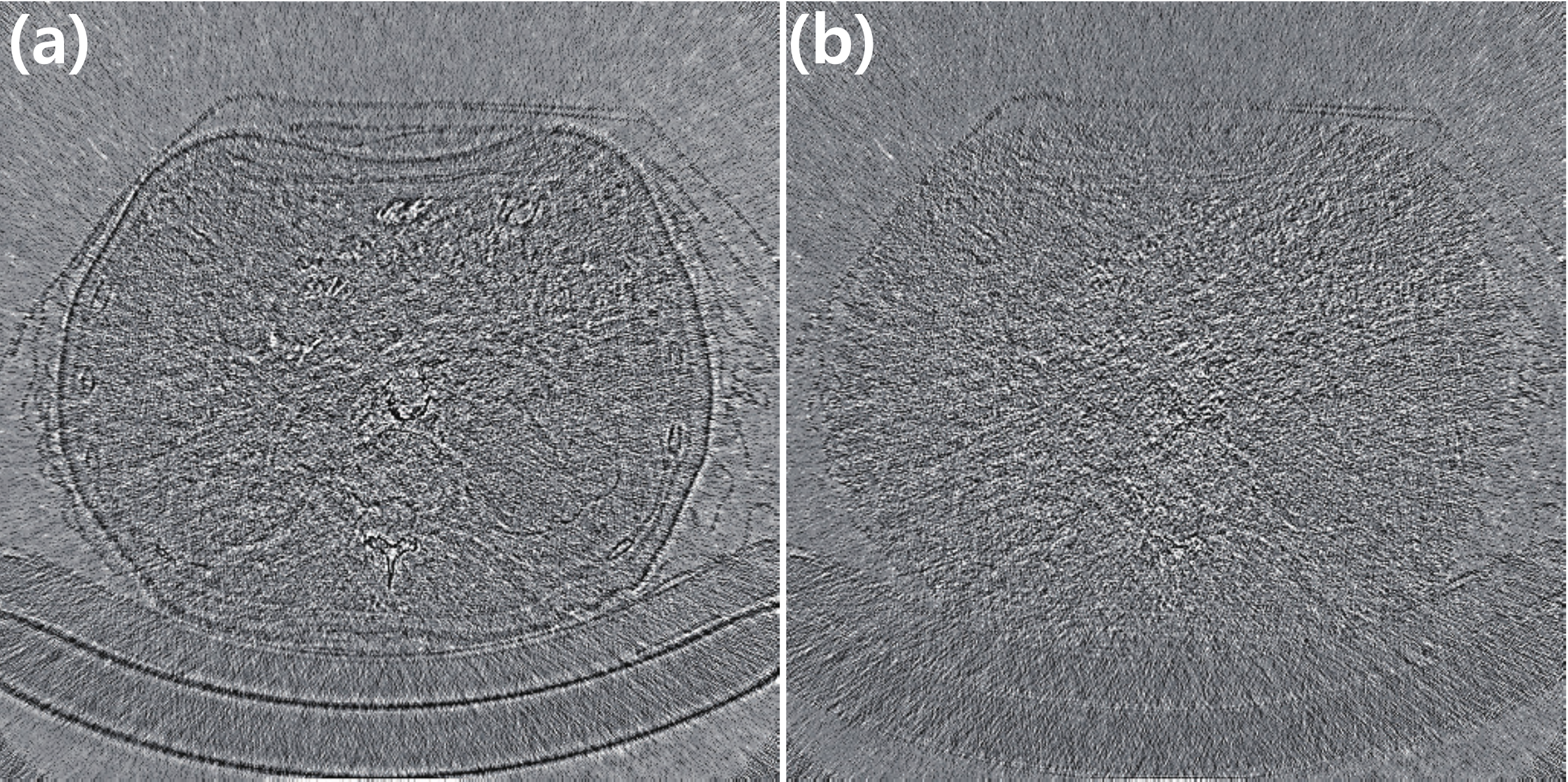}}
\caption{Difference images between the denoised images and routine dose image.
The intensity range was set to (-70,70) [HU].
(a) AAPM-Net, (b) WavResNet.}
\label{fig:difference_images}
\end{figure}

First, we presented the denoised images from the dataset of the one patient.
This  was excluded for the training with routine dose and quarter dose images in Fig. \ref{fig:result}(a)(b).
The result from AAPM-Net in Fig. \ref{fig:result}(c) significantly removes the low-dose noise, 
but the results appear  a bit blurry and lose some details of the texture.
On the other hand, the result of WavResNet in in Fig. \ref{fig:result}(d) clearly shows the significantly improved noise reduction,
while maintaining  edges and textures that help to distinguish the lesions in the organ.
In the lower part of Fig. \ref{fig:result}, the enlarged images are presented 
and the result of WavResNet clearly identifies the details like the vessels in the liver and some holes.
The difference images between denoised images and routine dose image in Fig. \ref{fig:difference_images}
 confirm the superiority of WavResNet.
The difference image of the proposed network contains only the noise of the low-dose X-ray CT image,
while  the difference image of AAPM-Net contains noise and the edge informations.

\begin{figure}[h]
\centering{\includegraphics[width=6cm]{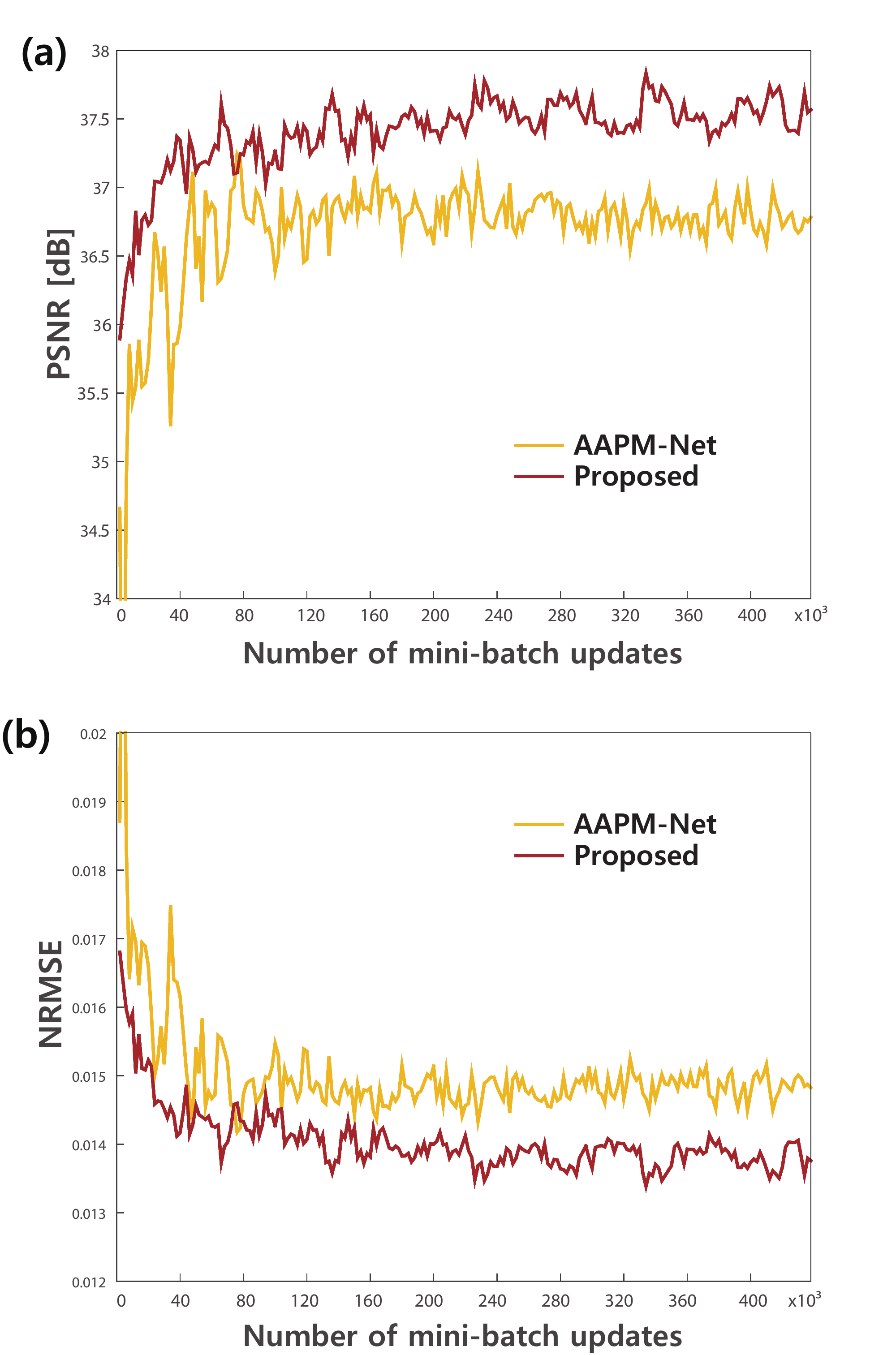}}
\caption{Convergence plots for (a) PSNR [dB] and (b) NRMSE with respect to each mini-batch updates.}
\label{fig:psnr_nrmse}
\end{figure}

The convergence plots in Fig. \ref{fig:psnr_nrmse} clearly shows the strength of WavResNet.
The proposed network exhibits the fast convergence and the final performance surpasses the AAPM-Net with respect to PSNR and NRMSE.

\subsection{Comparison with existing methods}

\begin{figure}[h]
\centering{\includegraphics[width=9cm]{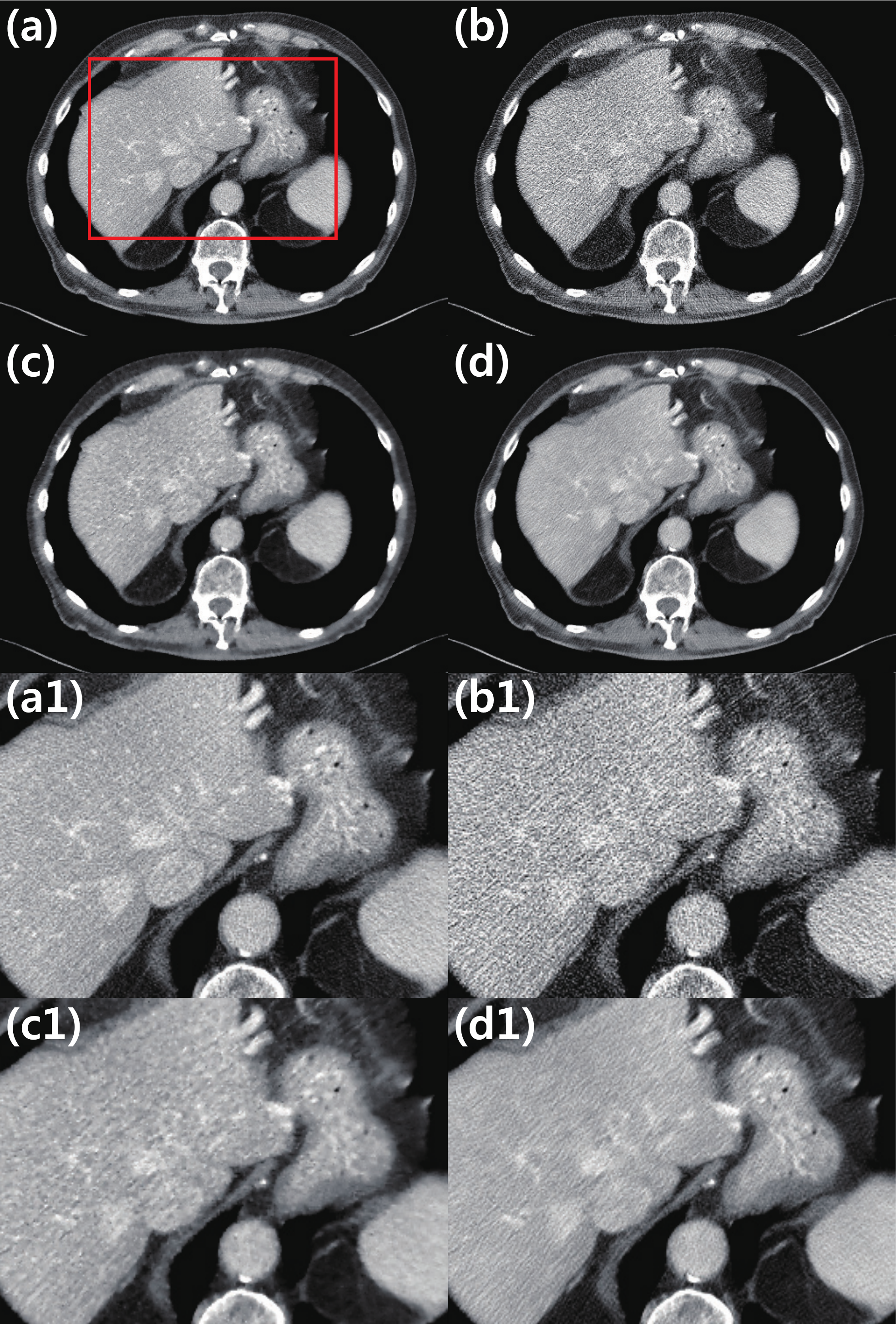}}
\caption{(a-d) Denoised images with routine dose and quarter dose images.
(a1-d1) Magnified images inside the red box.
The intensity range was set to (-160, 240) [HU] (Hounsfield Unit).
(a,a1) Routine dose, (b,b1) Quarter dose,
(c,c1) MBIR TV, (d,d1) WavResNet.}
\label{fig:result_mbir}
\end{figure}

\begin{figure}[h]
\centering{\includegraphics[width=9cm]{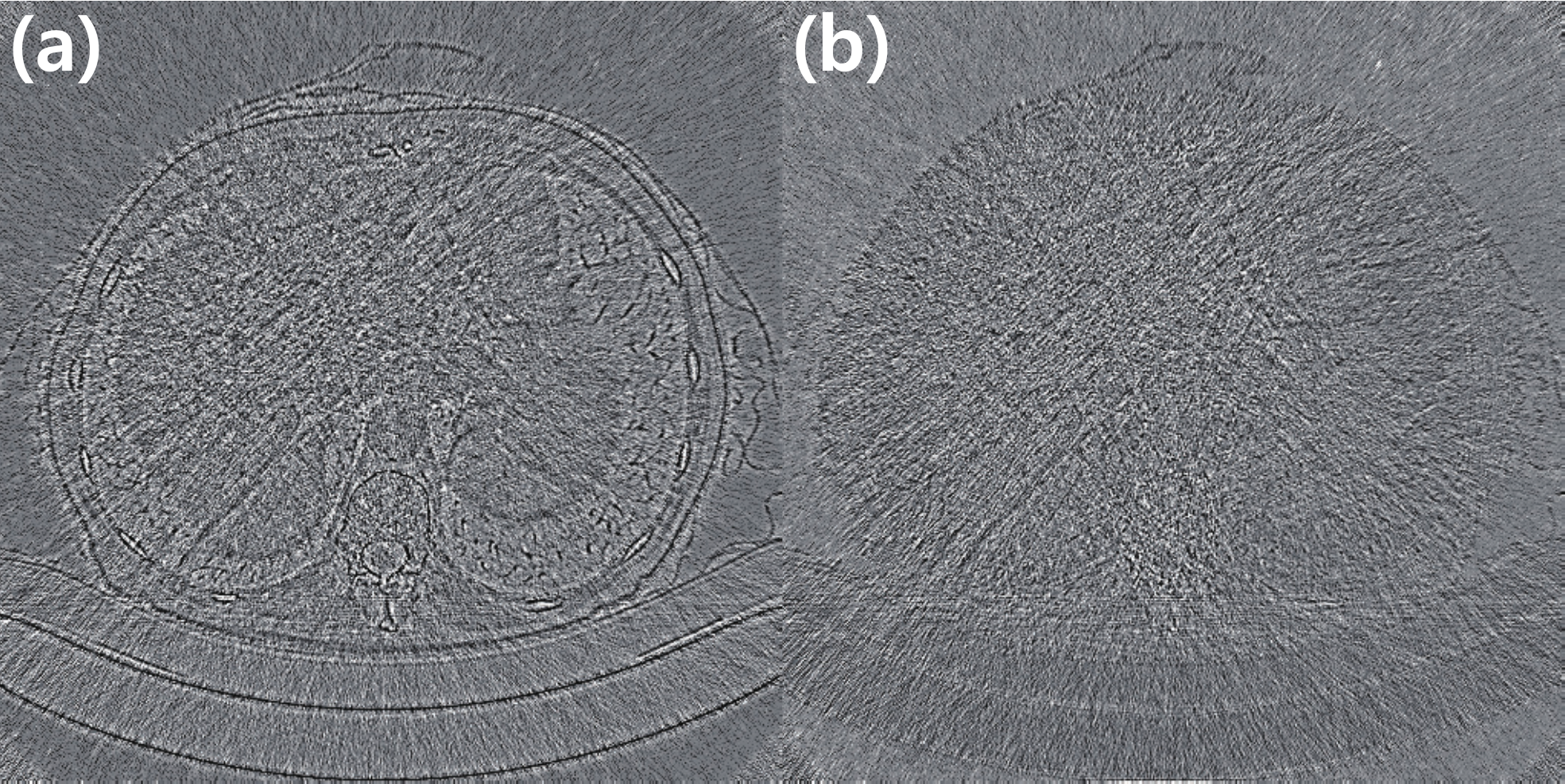}}
\caption{Difference images between the denoised images and routine dose image.
The intensity range was set to (-70,70) [HU].
(a) MBIR TV, (b) WavResNet.}
\label{fig:difference_mbir}
\end{figure}

Since MBIR is currently a standard low-dose reconstruction method,
we also compare the results of WavResNet and MBIR.
Fig.~\ref{fig:result_mbir} and Fig.~\ref{fig:difference_mbir} shows examples of comparative experiments.
The reconstruction results by MBIR appear a little blurred and the textures are reconstructed incorrectly.
On the other hand, WaveResNet provides clear reconstruction results.

Table \ref{table:psnr_nrmse} presents the averaged PSNR, NRMSE, and SSIM index values of the denoised images from 149 slices.
The proposed method showed the highest PSNR and SSIM index value and  has the lowest NRMSE value.

\begin{table}[h]
\centering
\begin{tabular}{l|c|c|c}
\hline \hline
 			& MBIR TV 	& AAPM-Net 	& Proposed 	 \\ 
\hline
PSNR [dB]  	& 35.34 	& 35.85 	& \bf{36.43} \\	
\hline
NRMSE 		& 0.017 	& 0.016 	& \bf{0.015} \\
\hline
SSIM index 	& 0.84 		& 0.85 		& \bf{0.87} \\
\hline \hline
\end{tabular}
\caption{Comparison of low-dose X-ray CT reconstruction algorithms.}
\label{table:psnr_nrmse}
\end{table}

\section{Conclusion}
\label{sec:conclusion}

In this paper, we proposed a wavelet domain  residual network  for low-dose X-ray CT reconstruction.
To improve the performance and retain the detailed textures,  a high-dimensional embedding scheme using 
contourlet and residual transform was proposed.
These processes simplify the topological structure of input and label data manifold, 
and  allows  fast convergence and improved performance.
The performance of WavResNet has been  verified by comprehensive comparative studies against
AAPM-Net and MBIR.

\section*{Acknowledgment}
\begin{small}
The authors would like to thanks Dr. Cynthia MaCollough,  the Mayo Clinic, the American Association of Physicists in Medicine (AAPM), 
and grant EB01705 and EB01785 from the National Institute of Biomedical Imaging and Bioengineering 
for providing the Low-Dose CT Grand Challenge data set.
This work is supported by Korea Science and Engineering Foundation, Grant number NRF-2016R1A2B3008104.
\end{small}


%

\end{document}